\documentclass[sigconf]{acmart}
\usepackage{enumitem}
\usepackage{array}
\usepackage{multirow}
\usepackage{longtable}
\usepackage{booktabs}
\usepackage{xcolor}
\usepackage{colortbl,booktabs}
\usepackage{color}
\usepackage{graphicx} 
\usepackage{subfigure}

\newcommand{\tabincell}[2]{\begin{tabular}{@{}#1@{}}#2\end{tabular}}

\AtBeginDocument{%
  }

\setcopyright{acmcopyright}
\copyrightyear{2023}
\acmYear{2023}
\setcopyright{acmlicensed}\acmConference[MM '23]{Proceedings of the 31st
ACM International Conference on Multimedia}{October 29-November 3,
2023}{Ottawa, ON, Canada}
\acmBooktitle{Proceedings of the 31st ACM International Conference on
Multimedia (MM '23), October 29-November 3, 2023, Ottawa, ON, Canada}
\acmPrice{15.00}
\acmDOI{10.1145/3581783.3612181}
\acmISBN{979-8-4007-0108-5/23/10}

\begin{document}

\title{Few-shot Multimodal Sentiment Analysis Based on Multimodal Probabilistic Fusion Prompts}

\author{Xiaocui Yang}
\authornote{This paper is completed at Singapore University of Technology and Design with funding from CSC.}
\affiliation{%
  \institution{Northeastern University, China}
  \country{}
}
\email{yangxiaocui@stumail.neu.edu.cn}

\author{Shi Feng}
\affiliation{%
  \institution{Northeastern University, China}
  \country{}
}
\email{fengshi@cse.neu.edu.cn}

\author{Daling Wang}
\affiliation{%
  \institution{Northeastern University, China}
    \country{}
}
\email{wangdaling@cse.neu.edu.cn}

\author{Yifei Zhang}
\affiliation{%
  \institution{Northeastern University, China}
    \country{}
}
\email{zhangyifei@cse.neu.edu.cn}

\author{Soujanya Poria}
\affiliation{%
  \institution{Singapore University of Technology and Design, Singapore}
  \country{}
}
\email{sporia@sutd.edu.sg}

\renewcommand{\shortauthors}{Xiaocui Yang et al.}

\begin{abstract}
 Multimodal sentiment analysis has gained significant attention due to the proliferation of multimodal content on social media. However, existing studies in this area rely heavily on large-scale supervised data, which is time-consuming and labor-intensive to collect. Thus, there is a need to address the challenge of few-shot multimodal sentiment analysis. To tackle this problem, we propose a novel method called \textbf{Multi}modal \textbf{P}r\textbf{o}babilistic Fus\textbf{i}o\textbf{n} Promp\textbf{t}s (\textbf{MultiPoint}\footnote{Our code and data can be found in the \url{https://github.com/YangXiaocui1215/MultiPoint}.}) that leverages diverse cues from different modalities for multimodal sentiment detection in the few-shot scenario.
 Specifically, we start by introducing a \textbf{C}onsistently \textbf{D}istributed \textbf{S}ampling approach called \textbf{CDS}, which ensures that the few-shot dataset has the same category distribution as the full dataset. 
 Unlike previous approaches primarily using prompts based on the text modality, we design unified multimodal prompts to reduce discrepancies between different modalities and dynamically incorporate multimodal demonstrations into the context of each multimodal instance. To enhance the model's robustness, we introduce a probabilistic fusion method to fuse output predictions from multiple diverse prompts for each input. 
 Our extensive experiments on \textbf{six} datasets demonstrate the effectiveness of our approach. First, our method outperforms strong baselines in the multimodal few-shot setting. Furthermore, under the same amount of data (1\% of the full dataset), our CDS-based experimental results significantly outperform those based on previously sampled datasets constructed from the same number of instances of each class.
\end{abstract}



\settopmatter{printacmref=false} 

\maketitle

\section{Introduction}
\label{sec:intro}
With the growing popularity of multimedia platforms, there has been an explosion of data containing multiple modalities such as text, image, video, and etc. Multimodal Sentiment Analysis (MSA) has emerged as a popular research topic due to its wide applications in market prediction, business analysis, and more  \cite{DBLP:journals/ijssmet/KaurK19, DBLP:journals/inffus/AbduYS21, DBLP:conf/icassp/ZhuZSN22}. In this paper, we specifically focus on the task of multimodal text-image sentiment analysis, which comprises of two subtasks: coarse-grained MSA and fine-grained MSA.
Coarse-grained MSA aims to detect the overall sentiment of a text-image pair \cite{DBLP:conf/sigir/XuMC18, DBLP:journals/tmm/YangFW021, DBLP:conf/acl/YangF0W20, DBLP:journals/corr/abs-2204-05515}. On the other hand, fine-grained MSA, also known as Multimodal Aspect-Based Sentiment Classification (MASC), seeks to detect the targeted sentiment for a specific aspect term that is dependent on the corresponding text-image pair \cite{DBLP:conf/acl/HuPHLL19, DBLP:conf/emnlp/JuZXLLZZ21, DBLP:conf/acl/LingYX22, DBLP:conf/emnlp/YangZ022, yu2022hierarchical, DBLP:journals/ipm/YangNY22}.
Multimodal sentiment analysis has witnessed significant progress in recent years. Early research primarily focuses on constructing rich and large-scale datasets to facilitate model training \cite{DBLP:conf/mmm/NiuZPE16, DBLP:journals/tmm/YangFW021, DBLP:conf/aaai/0001FLH18, DBLP:conf/acl/JiZCLN18, DBLP:journals/ijon/ZhouZHHH21}. Subsequent studies aim at improving the performance of MSA through the integration of various effective technologies, such as Contrastive Learning \cite{DBLP:journals/corr/abs-2204-05515}, 
Vision-Language Pre-training \cite{DBLP:conf/acl/LingYX22}, among others.


One of the limitations of existing multimodal sentiment analysis models is dependency on large-scale annotated datasets, which can be expensive and challenging to obtain. In real-world applications, only a limited amount of labeled data is available, making it more practical to investigate few-shot learning methods that can perform well in low-resource settings. 
However, in the multimodal few-shot learning setting, it can be challenging to sample diverse and comprehensive few-shot datasets. Existing few-shot classification tasks, such \cite{DBLP:conf/icmcs/YuZ22, DBLP:conf/mm/YuZL22}, typically sample the same number of instances for each label, without considering the consistency of the category distribution between the full dataset (before sampling) and the few-shot dataset (after sampling). This approach can result in imbalanced and biased few-shot datasets that do not reflect the true distribution of the full dataset.
To address this issue, we introduce a novel sampling approach called Consistently Distributed Sampling (\textbf{CDS}), which ensures that the few-shot dataset has a category distribution similar to that of the full dataset. 
\begin{figure*}[t] 
  \centering 
  \includegraphics[scale = 0.43]{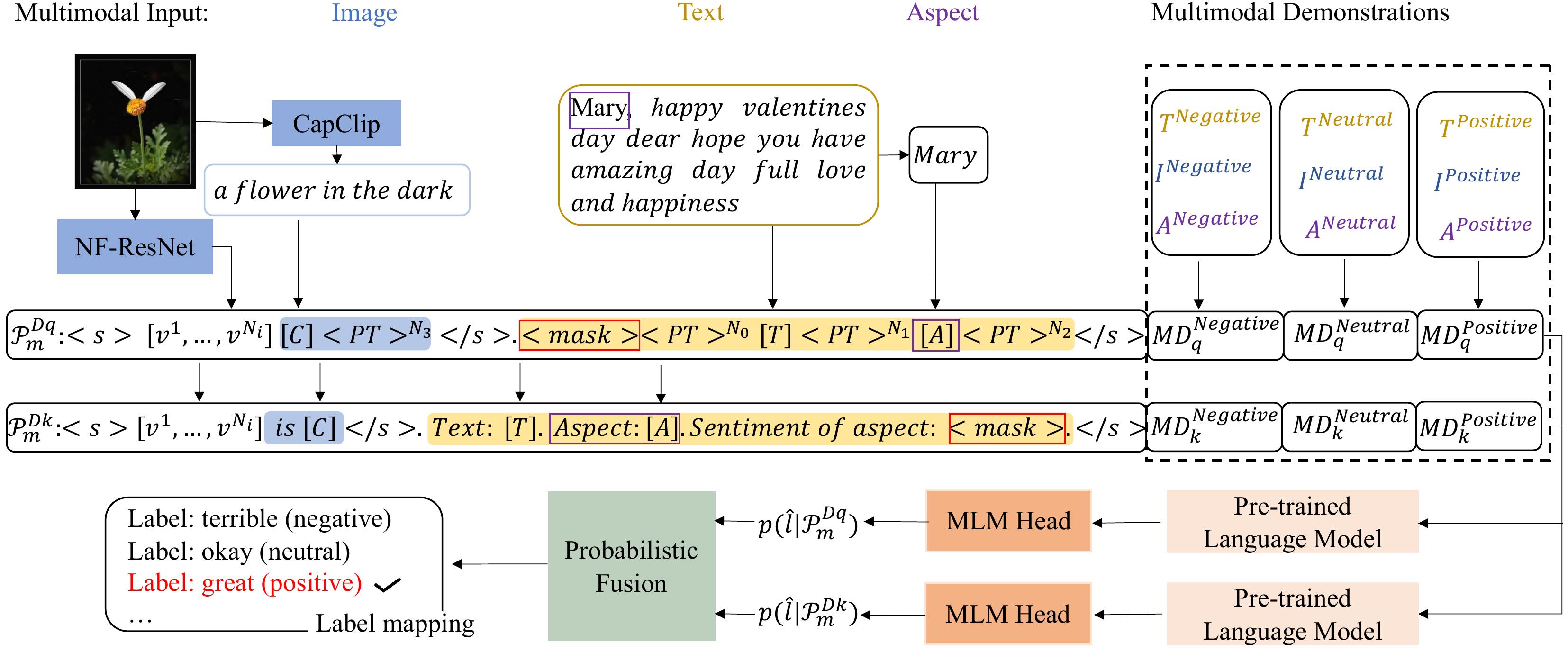} 
  \vspace{-1em}
  \caption{An illustration of our proposed Multimodal Probabilistic Fusion Prompts (MultiPoint) model for Few-shot Multimodal Sentiment Analysis. 
   We design different unified multimodal prompts with multimodal demonstrations, e.g., $\mathcal{P}_m^{Dq}$ and $\mathcal{P}_m^{Dk}$, here $q$ and $k$ indicate the q-th and k-th multimodal prompt for one instance.
   A multimodal prompt ($\mathcal{P}_m$) is composed of multiple image slots ([$v^1, ..., v^{N_i}$]), image prompt (blue highlight), and the task-specific text prompt (yellow highlight). $C$ is the image caption from ClipCap, $T$ is the original text, $A$ is an aspect term for fine-grained datasets, which does not exist in coarse-grained datasets. 
   $<mask>, <s>$ and $</s>$ are special tokens in Pre-trained Language Model. The black dashed boxes represent various demonstrations based on label space, take the $\mathcal{L} = \{Negative, Neutral, Positive\}$ as an example. The multimodal demonstration ${MD}_{q}$ for the q-th instance is dynamically selected based on the similarity score with the training dataset for a specific label from $\mathcal{L}$. Given a text-image pair, our model predicts the label $\hat{l}$.
   }
  \label{model_fig_1} 
\vspace{-1em}
\end{figure*}

Prompt-based methods have become popular in few-shot learning because they allow pre-trained models to generalize to new tasks with limited or no training data. Despite being widely used for few-shot text tasks, such as LM-BFF \cite{DBLP:conf/acl/GaoFC20} and GFSC \cite{DBLP:conf/naacl/Hosseini-AslLX22}, prompts are rarely utilized in multimodal scenarios.
To address this gap, Yu et al. propose a prompt-based vision-aware language modeling (PVLM) approach \cite{DBLP:conf/icmcs/YuZ22} and a unified pre-training for multimodal prompt-based fine-tuning (UP-MPF) \cite{DBLP:conf/mm/YuZL22} for multimodal sentiment analysis (MSA). 
PVLM and UP-MPF simply introduce image tokens to a pre-trained language model (PLM) for prompt-based fine-tuning.
However, directly feeding image representations into the language model raises the issue of modality discrepancy, as the image encoder is language-agnostic. This can result in suboptimal performance in capturing multimodal cues from multiple modalities.
Additionally, it has been observed that different prompts may contain varying amounts of information, and the information conveyed by a single prompt may be insufficient for effective multimodal sentiment analysis.
However, previous works on few-shot text tasks \cite{DBLP:conf/acl/GaoFC20, DBLP:conf/naacl/Hosseini-AslLX22} and multimodal tasks \cite{DBLP:conf/icmcs/YuZ22, DBLP:conf/mm/YuZL22} only apply a single prompt to different models, without considering the fusion of different prompts.

To alleviate the problems raised above, we propose a novel model for Few-shot MSA called \textbf{Multi}modal
\textbf{P}r\textbf{o}babilistic Fus\textbf{i}o\textbf{n} Promp\textbf{t}s, \textbf{MultiPoint}, depicted in Figure \ref{model_fig_1}.
To begin, we design unified multimodal prompts for our task, as shown in Table \ref{Templates}. For the text modality, we use both manual prompts based on domain knowledge and task-specific requirements, as well as generated prompts that capture diverse and valuable information from pre-trained language models. For the image modality, we generate a textual description of each image and use it as the image prompt to improve compatibility and mitigate discrepancies between the image and text modalities. The text and image prompts are then combined to create a unified multimodal prompt. 
To improve the robustness of our model, we select the most similar multimodal instances from the training dataset as multimodal demonstrations that are introduced as the multimodal context for each instance.
As previously mentioned, the information obtained from a single prompt is limited and different prompts can capture diverse cues from the data.
To this end, we propose a novel probabilistic fusion method, based on Bayesian Fusion, which has been shown to be robust in increasingly discrepant sub-posterior scenarios \cite{dai2021bayesian}. Our probabilistic fusion approach allows us to incorporate uncertainty in the predictions from different prompts and obtain a more reliable and accurate prediction for each instance.
We evaluate our approach on \textbf{six} multimodal sentiment datasets through extensive experiments.
Our main contributions are summarized as follows:
\vspace{-0.5em}
\begin{itemize} [leftmargin=*]
\item{
We  introduce a \textbf{C}onsistently \textbf{D}istributed \textbf{S}ampling approach, called \textbf{CDS}, which ensures that the category distribution of the few-shot dataset (only 1\% of the full dataset) is similar to that of the full dataset. This approach helps create representative few-shot datasets and enables more accurate evaluation of our model's performance.
}
\item{
We propose a novel model for Few-shot MSA called  \textbf{Multi}modal \textbf{P}r\textbf{o}babilistic Fus\textbf{i}o\textbf{n} Promp\textbf{t}s (\textbf{MultiPoint}). Our model employs unified multimodal prompts with multimodal demonstrations to mitigate the discrepancy between different modalities. Furthermore, probabilistic fusion aggregates predictions from multiple multimodal prompts, enhancing the effectiveness of our model.}
\item{We evaluate MultiPoint and CDS on \textbf{six} multimodal sentiment datasets. Our results in the few-shot setting demonstrate that MultiPoint outperforms strong baselines and showcase the benefits of utilizing consistent distribution information. }
\end{itemize}
\vspace{-1em}
\section{RELATED WORK}
\subsection{Multimodal Sentiment Analysis (MSA)}
MSA encompasses both coarse-grained MSA and fine-grained MSA.
\textbf{For coarse-grained MSA}, some datasets have been proposed include MVSA-Single and MVSA-Multiple datasets \cite{DBLP:conf/mmm/NiuZPE16}, and the TumEmo dataset \cite{DBLP:journals/tmm/YangFW021}.
Researchers have proposed various methods to tackle the challenges of multimodal sentiment analysis, including co-memory attentional model \cite{DBLP:conf/sigir/XuMC18},  Multi-channel Graph Neural Networks \cite{DBLP:conf/acl/YangF0W20}, Contrastive Learning and Multi-Layer Fusion (CLMLF) method \cite{DBLP:journals/corr/abs-2204-05515}, and more.
\textbf{For fine-grained MSA}, there are several datasets for aspect-based sentiment classification, including the Twitter-2015 and Twitter-2017 datasets \cite{DBLP:conf/aaai/0001FLH18, DBLP:conf/acl/JiZCLN18}. Additionally, a large-scale dataset called MASAD (Multimodal Aspect-based Sentiment Analysis Dataset) is built \cite{DBLP:journals/ijon/ZhouZHHH21} to facilitate research.
Several approaches have been proposed to address the challenges of fine-grained MSA. 
Initially, researchers expand BERT to the multimodal scenario, such as TomBERT \cite{DBLP:conf/ijcai/Yu019}, EF-CapTrBERT \cite{DBLP:conf/mm/0001F21}.
Recently, external knowledge is introduced to solve fine-grained MSA, e.g., FITE employing facial information \cite{DBLP:conf/emnlp/YangZ022}, KEF with knowledge-enhanced \cite{DBLP:conf/acl/LingYX22}, and VLP-MABSA leveraging external pre-training data and multiple pre-traing tasks \cite{DBLP:conf/acl/LingYX22}. 
Multimodal sentiment analysis has seen remarkable progress in recent years by leveraging fully supervised data or even other pre-trained data. 
However, collecting and annotating multimodal data for multimodal sentiment analysis is time-intensive and laborious. 
To this end, we devote to the multimodal sentiment analysis task in few-shot scenarios.
\vspace{-1.5em}
\subsection{Few-shot Learning with PLM}
In the field of Natural Language Processing (NLP), prompt-based language modeling has emerged as a powerful approach for solving different few-shot tasks using pre-trained language models (PLM) \cite{DBLP:journals/corr/abs-2107-13586}. Prompt-based methods treat the classification task as a masked language modeling (MLM) task, where the model is fine-tuned with a set of prompts to guide its predictions.
In the beginning, prompt-based approaches are introduced to handle text few-shot classification task, including LM-BFF \cite{DBLP:conf/acl/GaoFC20}, LM-SC \cite{DBLP:conf/naacl/JianGV22a}, and so on. 
Ehsan et al. \cite{DBLP:conf/naacl/Hosseini-AslLX22} propose a generative language model (GFSC) that reformulates the task as a language generation problem for text classification.
However, the above-mentioned models only handle text-related tasks. Recently, there has been an increasing interest in designing models to handle few-shot multimodal tasks. 
Existing models for few-shot multimodal tasks, such as Frozen \cite{DBLP:conf/nips/TsimpoukelliMCE21}, PVLM \cite{DBLP:conf/icmcs/YuZ22}, and UP-MPF \cite{DBLP:conf/mm/YuZL22}, primarily rely on introducing image tokens to a pre-trained language model for prompt-based fine-tuning. However, these approaches face the challenge of discrepancy between different modalities since image features are agnostic to language models.
To this end, we propose a novel unified multimodal prompt that allows for the joint processing of both text and image modalities in a coherent manner.
\vspace{-0.8em}
\section{Consistently Distributed Sampling}
\label{BFSD}
To construct the few-shot dataset for few-shot multimodal sentiment analysis task, it is important to select diverse samples that provide comprehensive coverage. 
Previous approaches \cite{DBLP:conf/icmcs/YuZ22, DBLP:conf/mm/YuZL22} have randomly sampled from the training and development sets to create few-shot datasets with equal amounts of data for each class, without taking into consideration 
the consistency of the category distribution between the full dataset (before sampling) and the few-shot dataset (after sampling).
Additionally, users express emotions with varying proportions on social media, indicating that the distribution of posts with different emotions are differ. 

We propose a novel sampling approach called Consistently Distributed Sampling (\textbf{CDS}). CDS ensures that the category distribution of the few-shot dataset is similar to that of the full dataset, creating representative few-shot datasets that reflect the real-world sentiment patterns observed on the internet. By constructing few-shot datasets using CDS, we can more accurately evaluate the performance of our model in a few-shot scenario. Specifically, we randomly sample about \textbf{1\%}\footnote{Following \cite{DBLP:conf/icmcs/YuZ22, DBLP:conf/mm/YuZL22}, we also randomly sample the 1\% data of training dataset as our few-shot training dataset.
} of the training dataset based on the sentiment distribution of the full training dataset as the few-shot multimodal training dataset, $\mathcal{D}_{train}$, and construct the development dataset, $\mathcal{D}_{dev}$, with the same sentiment distribution.
The MASAD dataset \cite{DBLP:journals/ijon/ZhouZHHH21} involves 57 aspect categories and 2 sentiments, and our sampled data considers the balance between different aspect categories and sentiments simultaneously.
For other datasests, we only consider balance of  sentiment categories.
The statistics of different datasets are given, as Table \ref{Dataset} and Table \ref{TumEmo} show.

\begin{table*}[t]
\begin{center}
\renewcommand{\arraystretch}{1.1} 
\caption{Unified multimodal templates for Few-shot Multimodal Sentiment Analysis. $\mathcal{P}$ is the template for the few-shot sentiment task, where $c$ represents coarse-grained datasets, $f$ represents fine-grained datasets, $t$ represents the text prompt, and $m$ represents the multimodal prompt. $T$ is the original text input, $\Tilde{V}$ is image slots from the image input $I$, and $A$ is the aspect term. The special tokens in the vocabulary of the pre-trained language model are represented as </s>, <mask>, and <PT>. The variable $n^p_{0,...,3}$ represents the number of learned prompt tokens, and for convenience, we set $n^p_0 = n^p_1 = n^p_2 = n^p_3$. Finally, there is a special token, <s>, at the front of each prompt, and the "$\oplus$" symbol denotes concatenation operation.
}
\vspace{-1em}
\begin{tabular}{
p{1cm}< \centering|
p{10cm}< \centering|
p{5.8cm}< \centering}
\toprule[1pt]
\textbf{Dataset} & \textbf{Text Prompts} &  \textbf{Unified Multimodal Prompts} \\
\cline{1-3}
\multirow{4}{*}{\textbf{\tabincell{c}{Coarse-\\grained}}}
& $\mathcal{P}_{t}^{c1} (T) $ = <s> [T] </s> It was <mask>.</s> & 
$\mathcal{P}_{m}^{c1} (T, I) $ =  <s> $\Tilde{V}$ is [C] </s> $\oplus$ $\mathcal{P}_{t}^{c1} (T) $  \\

& $\mathcal{P}_{t}^{c2} (T) $ =  <s> The sentence "[T]" has <mask> sentiment. </s> & 
$\mathcal{P}_{m}^{c2} (T, I) $ =  <s> $\Tilde{V}$ is [C] </s> $\oplus$ $\mathcal{P}_{t}^{c2} (T) $ \\

& $\mathcal{P}_{t}^{c3} (T) $ =  <s> Text: [T]. Sentiment of text: <mask>. </s> & 
$\mathcal{P}_{m}^{c3} (T, I) $ = <s>  $\Tilde{V}$ is [C] </s> $\oplus$  $\mathcal{P}_{t}^{c3} (T) $ \\

& $\mathcal{P}_{t}^{c4} (T) $ =  <s> <mask> <PT>$^{n^p_0}$ [T] <PT>$^{n^p_1}$ </s> & 
$\mathcal{P}_{m}^{c4} (T, I) $ =  <s> $\Tilde{V}$ [C]  <PT>$^{n^p_2}$ </s> $\oplus$  $\mathcal{P}_{t}^{c4} (T) $ \\
\midrule[1pt]
\multirow{4}{*}{\textbf{\tabincell{c}{Fine-\\grained}}}
& $\mathcal{P}_{t}^{f1} (T, A) $ = <s> [T] [A]</s> It was <mask>.</s> & 
$\mathcal{P}_{m}^{f1} (T, I, A) $ = <s> $\Tilde{V}$ is [C] </s> $\oplus$ $\mathcal{P}_{t}^{f1} (T, A) $  \\

& $\mathcal{P}_{t}^{f2} (T, A) $ = <s> The aspect "[A]" in sentence "[T]" has <mask> sentiment. </s> & 
$\mathcal{P}_{m}^{f2} (T, I, A) $ = <s> $\Tilde{V}$ is [C] </s> $\oplus$ $\mathcal{P}_{t}^{f2} (T, A) $ \\

& $\mathcal{P}_{t}^{f3} (T, A) $ = <s> Text: [T]. Aspect: [A]. Sentiment of aspect: <mask>. </s> & 
$\mathcal{P}_{m}^{f3} (T, I, A) $ = <s> $\Tilde{V}$ is [C] </s> $\oplus$ $\mathcal{P}_{t}^{f3} (T, A) $ \\

& $\mathcal{P}_{t}^{f4} (T, A) $ =  <s> <mask> <PT>$^{n^p_0}$ [T] <PT>$^{n^p_1}$  [A] <PT>$^{n^p_2}$ </s> & 
$\mathcal{P}_{m}^{f4} (T, I, A) $ =  <s> $\Tilde{V}$ [C] <PT>$^{n^p_3}$ </s> $\oplus$ $\mathcal{P}_{t}^{f4} (T, A) $ \\
\bottomrule[1pt]
\end{tabular}
\label{Templates}
\end{center}
\vspace{-1em}
\end{table*}
\section{Proposed Model}
\label{sec:method}
\subsection{Task Formulation}
We assume access to a pre-trained language model, denoted as $\mathcal{M}$, such as RoBERTa \cite{DBLP:journals/corr/abs-1907-11692}. Our goal is to fine-tune this model for the multimodal sentiment classification task on a specific label space, denoted as $\mathcal{L}$.
We construct $\mathcal{D}_{train}=\{(x^j)\}_{j=1}^{K}$ by CDS, where $K$ is the total number of text-image posts. Additionally, we choose the development set, $\mathcal{D}_{dev}$, to be the same size as the few-shot training set, i.e., $|\mathcal{D}_{dev}| = |\mathcal{D}_{train}|$.

\textbf{In Coarse-grained MSA,} $x^j = (t^j, i^j, l^j)$, where $t$ is the text modality, $i$ is the image modality, $l$ is the sentiment label for a text-image pair.
The model's objective is to predict the sentiment label $l$ for each text-image pair in an unseen test dataset $(t_{test}, i_{test}, l_{test}) \in \mathcal{D}_{test}$\footnote{For MVSA-Single and MVSA-Multiple, $l \in \{Negative, Neutral, Positive \}$. For TumEmo, $l \in \{Angry, Bored, Calm, Fear, Happy, Love, Sad \}$.}.

\textbf{In Fine-grained MSA,} $x^j = (t^j, i^j, a^j, l^j)$, where $t$ is the text modality, $i$ is the image modality, $a$ is the aspect term, $l$ is the sentiment label corresponding to the aspect term $a$.
The objective of the model is to predict the sentiment category $l$ for each aspect term based on the context of both the text and image modalities in the test dataset $(t_{test}, i_{test}, a_{test}, l_{test}) \ \in \mathcal{D}_{test}$\footnote{For Twitter-2015 and Twitter-2017, $l \in \{Negative, Neutral, Positive \}$; for MASAD,  $l \in \{Negative, Positive \}$.}.
\subsection{Multimodal Prompt-based Fine-tuning}
We propose a novel model called MultiPoint, which stands for Multi-modal Probabilistic Fusion Prompts. MultiPoint treats multimodal classification as a cloze-filling task, as depicted in Figure \ref{model_fig_1}. We first design separate prompts for different modalities and then create effective multimodal prompts for our task.
For the text modality, we manually design several text prompts, including $\mathcal{P}{t}^{c/f1}$, $\mathcal{P}{t}^{c/f2}$, and $\mathcal{P}{t}^{c/f3}$, and use the continuous text prompt, $\mathcal{P}{t}^{c/f4}$, to extract knowledge from PLMs. We believe that the manual prompts are carefully crafted based on domain knowledge and task-specific requirements, while the generated prompts are automatically generated from pre-trained language models to capture diverse and valuable information. The specific templates for the prompts are presented in Table \ref{Templates}.
For the image modality, $I$, we use ClipCap \cite{DBLP:journals/corr/abs-2111-09734} to generate a textual description of the image and use it as the image prompt, $C$, to bridge the gap between different modalities.
\begin{equation}
     C = ClipCap(I).
    \label{equ_2}
\end{equation}
We further leverage NF-ResNet \cite{DBLP:conf/iclr/BrockDS21} to extract and project the original image representation into the text feature space.
\begin{equation}
     V = W_i Pool(ResNet(I)) + b_i,
\end{equation}
\vspace{-1.5em}
\begin{equation}
\Tilde{V} = reshape(V)=[v^1, ..., v^j, ..., v^{N_i}], v^j \in \mathbb{R}^{d_t}, 
 \label{equ_3}
\end{equation}
where $V \in \mathbb{R}^{d_{nt}}$, $W_i \in \mathbb{R}^{d_v \times d_{nt}}$, $b_i \in \mathbb{R}^{d_{nt}}$. $nt = d_t \times N_i$, $N_i$, a hyperparameter, is the number of slots representing initial image representation in a multimodal prompt, and $d_t$ represents the dimension of text embedding in the pre-trained language model. 

Lastly, we design multiple multimodal prompts \textbf{$\mathcal{P}_m$} based on different text prompts, $\mathcal{P}_t$, and the image prompt. The specific unified multimodal prompts are presented in Table \ref{Templates}. We design three manual multimodal prompts, such as $\mathcal{P}^1_m$, $\mathcal{P}^2_m$, $\mathcal{P}^3_m$, as well as the continuous multimodal prompt \textbf{$\mathcal{P}^4_m$}. We choose to use only three manual prompts for demonstration purposes, as more similar prompts are also capable of handling MSA tasks in our actual experimental process.
\subsection{Multimodal Demonstrations}
Inspired by recent works, such as GPT-3 \cite{DBLP:conf/nips/BrownMRSKDNSSAA20} and LM-BFF \cite{DBLP:conf/acl/GaoFC20}, we further design multimodal demonstrations chosen by similarity scores, as shown on the right side of Figure \ref{model_fig_1}. Specifically, we first feed the raw text input $t$ and image prompt $c$ from the image input $i$, that can be regarded as text description of image,  into a pre-trained language model, such as SBERT \cite{DBLP:conf/emnlp/ReimersG19}, to obtain embeddings $E$.
\begin{equation}
    E = SBERT([t \oplus a \oplus c]),
    \label{equ_sbert}
\end{equation}
where $\oplus$ is the concatenation operation. $a$ represents the aspect term and is optional. For the fine-grained task, we combine the text with the aspect term, while for the coarse-grained task, there is no aspect term.

Next, we compute the similarity scores between each query instance $x_{que} = (t_{que}, i_{que}, a_{que})$ and support set with $K^l$ instances for the $l$-th label category, $D^{(l)}_{sup} = {\{{(x_{sup}^{(l)})}^j\}}_{j=1}^{K^l}$. It is worth noting that the support instances are taken from the training dataset $\mathcal{D}_{train}$, both during training and inference stages.
\begin{equation}
    \label{equ_similarity_score1}
    Sim(x_{que}, x^{(l)}_{sup}) = cos (E_{que}, E^{(l)}_{sup}).
\end{equation}

We then select the multimodal support instance with the highest similarity score for each label category $l$.
\begin{equation}
    \label{equ_similarity_score2}
    x_{sup}^{best^{(l)}} = \mathop{\arg\max}\limits_{Label=l, j} Sim(x_{que}, x^j_{sup})_{j=1}^{K^l}.
\end{equation}

Finally, we convert the multimodal support instances with the highest similarity scores into $\mathcal{P}_m$ templates, with <mask> tokens replaced by different labels from $\mathcal{L}$. These resulting multimodal prompts are denoted as $\hat{\mathcal{P}}_m$, and we concatenate them with the query instance $x_{que}$.
\begin{equation}
    \label{equ_similarity_score3}
    \mathcal{P}_m^D =  \mathcal{P}_m(x_{que}) \oplus \hat{\mathcal{P}}_m(x_{sup}^{best^{(1)}}, l^{(1)}) \oplus ... \oplus \hat{\mathcal{P}}_m(x_{sup}^{best^{(|\mathcal{L}|)}}, l^{(|\mathcal{L}|)}),
\end{equation}
where $|\mathcal{L}|$ is the number of sentiment categories in each dataset. 
\subsection{Classification}

Let $\boldsymbol{\phi}:\mathcal{L} \rightarrow \mathcal{V}$ be a mapping from the task label space to individual words in the vocabulary $\mathcal{V}$ of the pre-trained language model, $\mathcal{M}$.
For each text-image pair $x = (t, i)$ for a coarse-grained dataset or $x = (t, i, a)$ for a fine-grained dataset, we input the multimodal prompt from Eq. \ref{equ_similarity_score3}, $\mathcal{P}^{D}_m$, that contains the $<mask>$ token into the MLM head. We cast our multimodal classification task as a cloze problem and model the probability of predicting class $\hat{l} \in \mathcal{L}$ as:
\begin{equation}
\begin{aligned}
        p(\hat{l} | \mathcal{P}^D_m(x)) & = p(<mask> = \boldsymbol{\phi}( \hat{l}) | \mathcal{P}^D_m) \\
                       & = \frac{exp(\textbf{w}_{\boldsymbol{\phi}(\hat{l})} \cdot \textbf{h}_{<mask>})}
                       {\sum_{l^{'} \in \mathcal{L}} exp(\textbf{w}_{\boldsymbol{\phi}(l^{'})} \cdot \textbf{h}_{<mask>})} ,
    \label{equ_6}
\end{aligned}
\end{equation}
where $\textbf{h}_{<mask>}$ is the hidden representation of <mask> token and $\textbf{w}_v$ indicates the final layer weight of MLM corresponding to $v \in \mathcal{V}$.
\subsection{Multimodal Probabilistic Fusion}
We find that  different prompts contain various amounts of information, and the information conveyed by a single prompt is insufficient.
We fuse prediction logits from different multimodal prompts based on Bayes Rule \cite{chen2021multimodal, dai2021bayesian}  to provide more robust detection than a single prompt. 
For instance, there are $n$ multimodal prompts $ \{\mathcal{P}_m^{D1}, ..., \mathcal{P}_m^{Dn} \}$.
Crucially, given  one instance $x$  that label is classified as $\hat{l}$ by $\mathcal{M}$, we assume that different multimodal prompts are conditionally independent.
\begin{equation}
    p(\mathcal{P}_m^{D1}, ..., \mathcal{P}_m^{Dn} | \hat{l} ) =p(\mathcal{P}_m^{D1} | \hat{l} ) ... p(\mathcal{P}_m^{Dn} | \hat{l} ) .
    \label{equ_7}
  \end{equation}
Therefore, assuming conditional independence between the prediction results of the MLM for different multimodal prompts, we perform multimodal sentiment detection using multiple prompts and propose a novel multimodal probabilistic fusion approach.
\begin{equation}
\begin{aligned}
    p(\hat{l} | \mathcal{P}_m^{D1}, ..., \mathcal{P}_m^{Dn})  & = \frac{p(\mathcal{P}_m^{D1}, ...,  \mathcal{P}_m^{Dn} | \hat{l} ) p(\hat{l})}{p(\mathcal{P}_m^{D1}, ..., \mathcal{P}_m^{Dn})} \\
                            & \propto p(\mathcal{P}_m^{D1}, ..., \mathcal{P}_m^{Dn} | \hat{l} ) p(\hat{l} ) \\
                            & \propto p(\mathcal{P}_m^{D1} | \hat{l}) ... p(\mathcal{P}_m^{Dn} | \hat{l} ) p(\hat{l} ) \\
                            & \propto \frac{p(\mathcal{P}_m^{D1} | \hat{l}) p(\hat{l} ) ...  p(\mathcal{P}_m^{Dn} | \hat{l} ) p(\hat{l}) p(\hat{l})}{{p(\hat{l} )}^n}  \\ & \propto \frac{p(\hat{l}  | \mathcal{P}_m^{D1}) ... p(\hat{l} | \mathcal{P}_m^{Dn})}{{p(\hat{l})}^{n-1}} . 
    \label{equ_8}
\end{aligned}
\end{equation}
We first train independent classifiers that predict the distributions over the label $\hat{l}$ given each individual multimodal prompt, such as $p(\hat{l} | \mathcal{P}_m^{Dk})$. Then, we obtain the fused distribution of label $\hat{l}$ from the $n$ multimodal prompts based on the probabilistic fusion module.
\begin{equation}
    p(\hat{l}  | \{\mathcal{P}_m^{Dk}\}_{k=1}^n) \propto \frac{\prod_{k=1}^n p(\hat{l}  | \mathcal{P}_m^{Dk})}{{p(\hat{l})}^{n-1}} ,
    \label{equ_9}
\end{equation}
where we set $n=2$ due to computational resource constraints, which is sufficient to demonstrate the effectiveness of our approach.
\section{EXPERIMENTS}
\subsection{Datasets}
We evaluate our proposed model on six multimodal sentiment datasets, including three coarse-grained datasets (MVSA-Single, MVSA-Multiple, and TumEmo) and three fine-grained datasets (Twitter-2015, Twitter-2017, and MASAD), where the label sets $\mathcal{L}$ vary across different datasets. Following \cite{DBLP:conf/icmcs/YuZ22}, we keep the test set unchanged and sample data based on CDS to form few-shot datasets, consisting of about 1\% of the training set with $K_{train} = K_{dev}$. The statistics of the different datasets are presented in Tables \ref{Dataset} and \ref{TumEmo}. The specific method of sampling data is described in Section \ref{BFSD}.

\begin{table*}[t] \small
\begin{center}
\renewcommand{\arraystretch}{1} 
\caption{Statistics for five datasets, including MVSA-Single, MVSA-Multiple, Twitter-2015, Twitter-2017, and MASAD. For A/B, B represents the number of original data, and A represents the number of few-shot data sampled based on CDS.
 For all datasets, the few-shot dataset represents approximately 1\% of the overall training data. In the few-shot setting, the number of development datasets is equal to the number of training datasets.
}
\vspace{-1em}
\begin{tabular}{
p{2.2cm}< \centering p{2.2cm}< \centering|
p{1.1cm}< \centering p{1.1cm}< \centering p{1.1cm}< \centering p{1.2cm}< \centering|
p{1.1cm}< \centering p{1.1cm}< \centering p{1.1cm}< \centering p{1.2cm}< \centering}
\toprule[1pt]
\multicolumn{2}{p{4.4cm}<\centering|}{\multirow{2}{*}{\textbf{Dataset}}} &
\multicolumn{4}{c|}{\textbf{Train}} & 
\multicolumn{4}{c}{\textbf{Test}} \\
\cline{3-10}
& & \textbf{Negative} & \textbf{Neutral} &\textbf{Positive} & \textbf{Total} & \textbf{Negative} & \textbf{Neutral} &\textbf{Positive} & \textbf{Total} \\
\midrule[1pt]
\multirow{2}{*}{\textbf{Coarse-grained}}
& MVSA-Single & \textbf{10}/1004 & \textbf{4}/345 & \textbf{20}/1921 & \textbf{34}/3270
& 126 & 37 & 249 & 412\\
& MVSA-Multiple & \textbf{20}/1909 & \textbf{32}/3170 & \textbf{82}/8166 & \textbf{134}/13245 & 217 & 405 & 1014 & 1636  \\
\midrule[1pt]
\multirow{3}{*}{\textbf{Fine-grained}}
& Twitter-2015 & \textbf{4}/368 & \textbf{19}/1883 & \textbf{10}/928 & \textbf{33}/3179 
& 113 & 607 & 317 & 1037 \\
& Twitter-2017 & \textbf{4}/416 & \textbf{16}/1638 & \textbf{15}/1508 & \textbf{35}/3562
& 168 & 573 & 493 & 1234 \\
& MASAD & \textbf{69}/5605 & \textbf{0}/0 & \textbf{101}/9263 & \textbf{170}/14868
& 1767 & 0 & 3168 & 4935\\
\bottomrule[1pt]
\end{tabular}
\label{Dataset}
\end{center}
\vspace{-1em}
\end{table*}

\begin{table*}[h] 
\begin{center}
\renewcommand{\arraystretch}{1} 
\caption{Statistics for the TumEmo dataset that has the same few-shot setting as other datasets.
}
\vspace{-1em}
\begin{tabular}{
p{2.2cm}< \centering|
p{1.4cm}< \centering p{1.4cm}< \centering p{1.4cm}< \centering p{1.4cm}< \centering
p{1.4cm}< \centering p{1.4cm}< \centering p{1.4cm}< \centering p{1.4cm}< \centering}
\toprule[1pt]
\textbf{Dataset} & \textbf{Angry} & \textbf{Bored} & \textbf{Calm} & \textbf{Fear} & \textbf{Happy} & \textbf{Love} & \textbf{Sad} & \textbf{Total} \\
\cline{1-9}
\textbf{Train} & \textbf{60}/5879 & \textbf{108}/10823 & \textbf{63}/6300 & \textbf{86}/8625 & \textbf{222}/22215 & \textbf{150}/15016 & \textbf{68}/6829 & \textbf{757}/75687 \\
\textbf{Test} &  736 & 1354 & 788 & 1079 & 2776 & 1875 & 855 & 9463 \\
\bottomrule[1pt]
\end{tabular}
\label{TumEmo}
\end{center}
\vspace{-1em}
\end{table*}

\subsection{Experimental Setup}
In the text prompt, we use the original label set for TumEmo, which has multiple emotion labels. For other datasets, we map the label set \{negative, neutral, positive\} to \{terrible, okay, great\}.
Our model is constructed using RoBERTa-large with 355M parameters, $\mathcal{M}$.
Fine-tuning on small datasets can suffer from instability, and results may change dramatically given a new data split \cite{DBLP:conf/iclr/0007WKWA21, DBLP:conf/acl/GaoFC20}. To account for this, we measure average performance across five randomly sampled $\mathcal{D}_{train}$ and $\mathcal{D}_{dev}$ splits based on different seeds, i.e., \textit{13, 21, 42, 87, 100}. To provide a more reliable measure of performance, we repeat the experiment three times for each split, resulting in a total of 15 ($3 \times 5$) training runs for each dataset. We report the mean Accuracy (Acc), Weighted-F1 (F1)\footnote{Since most datasets have highly imbalanced categories, the Weighted-F1 value is a more reasonable metric.}, and the standard deviation over the 15 runs.
We set the batch-size to 8. For the number of prompt tokens for $\mathcal{P}^4$ in Table \ref{Templates}, we set $n^p_0 = n^p_1 = n^p_2 = n^p_3 = 1$ for Twitter-2017 and MASAD and $n^p_0 = n^p_1 = n^p_2 = n^p_3 = 2$ for other datasets.
Our model performs best in the Acc metric when $N^i =1$ in Eq. \ref{equ_3}, and we set learning rates of 5e-6/2e-6/1e-5/3e-6 for MVSA-Single/Twitter-2017/MASAD/other datasets. Unless otherwise specified, we use these hyperparameters.
MultiPoint has a total of approximately 410M parameters, and all parameters are updated during training. The training time varies depending on the dataset. For example, we train our model up to 1000 training steps in approximately 60 minutes for the MVSA-Single,  MVSA-Multiple, Twitter-2015, and Twitter-2017 datasets. For the MASAD/TumEmo dataset, training for 1000 training steps takes around 100/120 minutes.

\subsection{Baselines}
We compare our model with three groups of baselines\footnote{Unless otherwise specified, all baselines are based on RoBERTa-large.}. \textbf{The first group} consists of previous text-based models, including \textbf{RoBERTa} \cite{DBLP:journals/corr/abs-1907-11692}, \textbf{Prompt Tuning (PT)} only uses a single textual prompt based on the multimodal prompt, such as [<s> [T] \ It \ was \ <mask>. \\</s>] for coarse-grained datasets and [<s> [T] [A] \ It \ was \ <mask>.\\</s>] for fine-grained datasets, \textbf{LM-BFF} \cite{DBLP:conf/acl/GaoFC20} utilizes generated text prompts based on each specific dataset and text demonstrations to solve few-shot text classification tasks, \textbf{LM-SC} \cite{DBLP:conf/naacl/JianGV22a} introduces supervised contrastive learning based on LM-BFF to few-shot text tasks, and \textbf{GFSC} \cite{DBLP:conf/naacl/Hosseini-AslLX22} converts the classification task into a generation task to solve text classification tasks in the few-shot setting through the pre-trained generation model, i.e., GPT2 \cite{radford2018improving}.

\textbf{The second group }consists of multimodal approaches that are trained in full MSA datasets from published papers. \textbf{For the coarse-grained MSA task:}
\textbf{Multimodal Fine Tuning (MFN)} is a baseline that doesn't use any designed prompts and employs the representation of the ``<s>'' token for classification.
\textbf{CLMLF} \cite{DBLP:journals/corr/abs-2204-05515} is the state-of-the-art model for coarse-grained MSA.
\textbf{For the fine-grained MSA task:}
\textbf{TomBERT} \cite{DBLP:conf/ijcai/Yu019} is a multimodal BERT for the fine-grained MSA task.
\textbf{EF-CapTrBERT} \cite{DBLP:conf/mm/0001F21} translates images in input space to construct an auxiliary sentence that provides multimodal information to BERT.
\textbf{KEF} \cite{DBLP:conf/acl/LingYX22} exploits adjective-noun pairs extracted from the image for the fine-grained MSA task.
\textbf{FITE} \cite{DBLP:conf/emnlp/YangZ022} is the state-of-the-art model for fine-grained MSA, which leverages facial information from the image modality.
\textbf{VLP-MABSA} \cite{DBLP:conf/acl/LingYX22} designs a unified multimodal encoder-decoder architecture and different pre-training tasks to improve the fine-grained MSA task.

\textbf{The last group} includes multimodal approaches that have been trained for few-shot MSA.
\textbf{PVLM} \cite{DBLP:conf/icmcs/YuZ22} directly introduces image features to pre-trained language models to solve the MAS task in a few-shot scenario.
\textbf{UP-MPF} \cite{DBLP:conf/mm/YuZL22} is the state-of-the-art model in the multimodal few-shot setting for the MSA task. It further employs pre-training data and tasks based on PVLM.
\textbf{MultiPoint} is our model that introduces multiple multimodal prompts with demonstrations and probabilistic fusion to improve the performance of MSA in a few-shot scenario.
Note that we reproduced the LM-BFF, LM-SC, EF-CapTrBERT, FITE, VLP-MABSA, PVLM, and UP-MPF models based on the RoBERTa-large model, while TomBERT and KEF are based on the BERT-base model.

\begin{table*}[t] \small
\begin{center}
\caption{Our main results for few-shot experiments on three multimodal coarse-grained datasets, including MVSA-Single, MVSA-Multiple, and TumEmo. The standard deviation is in parentheses.
``$*$'' indicates baselines with prompt tuning and applies multiple prompts from Table \ref{Templates}. We report the best performance of the baselines applying different prompts. 
``$\mathcal{P}_{m}$'' means the multimodal prompt, $c$ is coarse-grained. ``[q-k]'' means combine q-th prompt with k-th prompt.
}
\vspace{-1em}
\renewcommand{\arraystretch}{1} 
\begin{tabular}{
p{1.6cm}< \centering|
p{3cm}< \centering|
p{1.6cm}< \centering p{1.6cm}< \centering|
p{1.6cm}< \centering p{1.6cm}< \centering|
p{1.6cm}< \centering p{1.6cm}< \centering}
\toprule[1pt]
\multirow{2}{*}{\textbf{Modality}} &
\multirow{2}{*}{\textbf{Model}} & 
\multicolumn{2}{p{3.2cm}<\centering|}{\textbf{MVSA-Single}} & \multicolumn{2}{p{3.2cm}<\centering|}{\textbf{MVSA-Multiple}} & \multicolumn{2}{p{3.2cm}<\centering}{\textbf{TumEmo}}\\
\cline{3-8}
& & \textbf{Acc} & \textbf{F1} &\textbf{Acc} & \textbf{F1} & \textbf{Acc} & \textbf{F1} \\
\midrule[1pt]
\multirow{5}{*}{\textbf{Text}}
& RoBERTa & 61.21\,($\pm$2.11) & 56.11\,($\pm$2.74) & 63.40\,($\pm$0.86) & 61.34\,($\pm$1.40) & 55.03\,($\pm$0.45) & 54.87\,($\pm$0.57)\\
& PT$^{*}$  & 65.73\,($\pm$1.96) & 64.13\,($\pm$1.77) & 65.91\,($\pm$1.88) & 64.05\,($\pm$1.42) & 55.97\,($\pm$0.30) & 55.84\,($\pm$0.33)\\
& LM-BFF$^{*}$  & 65.58\,($\pm$2.81) & 63.41\,($\pm$3.00) & 66.36\,($\pm$0.88) & 64.08\,($\pm$1.09) & 56.03\,($\pm$0.66) & 55.85\,($\pm$0.63)\\
& LM-SC$^{*}$  & 66.51\,($\pm$1.09) & 64.62\,($\pm$0.98) & 65.37\,($\pm$0.87) & 63.63\,($\pm$1.56) & 55.95\,($\pm$0.40) & 56.00\,($\pm$0.52)\\
& GFSC$^{*}$ & 63.39\,($\pm$4.10) & 58.72\,($\pm$6.52) & 64.72\,($\pm$1.18) & 63.53\,($\pm$0.56) & 53.28\,($\pm$0.50) & 52.83\,($\pm$0.59)\\
\midrule[1pt]
\multirow{2}{*}{\textbf{\tabincell{c}{Text-Image}}}
& MFN   & 64.08\,($\pm$2.44) & 60.60\,($\pm$2.97) & 64.04\,($\pm$1.97) & 61.46\,($\pm$1.98) & 56.83\,($\pm$0.38) & 56.82\,($\pm$0.40)\\
& CLMLF  & 61.19\,($\pm$0.65) & 51.34\,($\pm$3.24) & 63.86\,($\pm$1.76) & 57.95\,($\pm$4.32) & 42.65\,($\pm$9.32) & 38.41\,($\pm$12.19)\\
\midrule[1pt]
\multirow{8}{*}{\textbf{\tabincell{c}{Text-Image}}}
& PVLM$^{*}$     & 66.94\,($\pm$1.20) & 63.10\,($\pm$2.79) & 67.40\,($\pm$0.99) & 63.67\,($\pm$2.56) & 55.43\,($\pm$0.72) & 55.02\,($\pm$0.70)\\
& UP-MPF$^{*}$    & 66.84\,($\pm$2.05) & 64.96\,($\pm$1.37) & 67.35\,($\pm$0.97) & 61.00\,($\pm$2.23) & 54.91\,($\pm$0.94) & 54.38\,($\pm$1.05)\\
& MultiPoint(${\mathcal{P}^{c[1-4]}_{m}}$)   & \textbf{69.95\,($\pm$2.47)} & \textbf{68.60\,($\pm$1.73)} & 68.04\,($\pm$0.57) & 65.39\,($\pm$1.28)  & \textbf{58.09\,($\pm$0.43)} & 58.05\,($\pm$0.37) \\
& MultiPoint(${\mathcal{P}^{c[2-4]}_{m}}$)   & 69.66\,($\pm$1.48) & 67.96\,($\pm$1.13) &  67.67\,($\pm$0.85) & 65.15\,($\pm$1.47) & 57.97\,($\pm$0.51) & 57.92\,($\pm$0.47)\\
& MultiPoint(${\mathcal{P}^{c[3-4]}_{m}}$)   & 69.76\,($\pm$1.08) & 68.02\,($\pm$1.47) & \textbf{68.27\,($\pm$1.15)} & \textbf{65.34\,($\pm$1.87)} & 58.05\,($\pm$0.53) & \textbf{58.06\,($\pm$0.50)}\\
& MultiPoint(${\mathcal{P}^{c[1-2]}_{m}}$) & 68.11\,($\pm$1.40) & 67.03\,($\pm$1.05) & 67.24\,($\pm$0.87) & 64.83\,($\pm$1.34) & 57.74\,($\pm$0.45) &  57.69\,($\pm$0.45) \\
& MultiPoint(${\mathcal{P}^{c[1-3]}_{m}}$) & 68.59\,($\pm$0.59) & 67.40\,($\pm$0.88) & 67.63\,($\pm$1.15) & 65.28\,($\pm$1.32) & 57.80\,($\pm$0.77) & 57.79\,($\pm$0.71)\\ 
& MultiPoint(${\mathcal{P}^{c[2-3]}_{m}}$) & 68.15\,($\pm$1.98) & 66.87\,($\pm$1.39) & 67.12\,($\pm$1.43) & 65.04\,($\pm$1.29) & 57.48\,($\pm$0.68) & 57.44\,($\pm$0.64) \\
\bottomrule[1pt]
\end{tabular}
\label{coarse-grained results}
\end{center}
\end{table*}

\begin{table*}[t] \small
\begin{center}
\caption{Our main results for few-shot experiments on three multimodal fine-grained datasets, including Twitter-2015, Twitter-2017, and MASAD. The standard deviation is in parentheses. $f$ represents fine-grained. ``--'' indicates no reproducible results on MASAD, as these baselines require external knowledge to model, such as captions, adjective-noun pairs, etc.}
\vspace{-1em}
\renewcommand{\arraystretch}{1} 
\begin{tabular}{
p{1.6cm}< \centering|
p{3cm}< \centering|
p{1.6cm}< \centering p{1.6cm}< \centering|
p{1.6cm}< \centering p{1.6cm}< \centering|
p{1.6cm}< \centering p{1.6cm}< \centering}
\toprule[1pt]
\multirow{2}{*}{\textbf{Modality}} &
\multirow{2}{*}{\textbf{Model}} & 
\multicolumn{2}{p{3.2cm}<\centering|}{\textbf{Twitter-2015}} & \multicolumn{2}{p{3.2cm}<\centering|}{\textbf{Twitter-2017}} & \multicolumn{2}{p{3.2cm}<\centering}{\textbf{MASAD}}\\
\cline{3-8}
& & \textbf{Acc} & \textbf{F1} &\textbf{Acc} & \textbf{F1} & \textbf{Acc} & \textbf{F1} \\
\midrule[1pt]
\multirow{5}{*}{\textbf{Text}}
& RoBERTa & 55.58\,($\pm$4.13) & 52.32\,($\pm$2.28) & 48.22\,($\pm$2.95) & 46.37\,($\pm$3.17) & 68.81\,($\pm$1.76) & 67.88\,($\pm$1.43)\\
& PT$^*$ & 61.97\,($\pm$3.15) & 60.11\,($\pm$3.38) & 58.77\,($\pm$3.70) & 57.85\,($\pm$3.63) & 77.62\,($\pm$1.34) & 77.60\,($\pm$1.37)\\
& LM-BFF$^{*}$ & 60.87\,($\pm$3.38) & 59.63\,($\pm$3.04) & 56.84\,($\pm$3.51) & 55.96\,($\pm$3.48) & 78.87\,($\pm$0.94) & 78.35\,($\pm$0.77)\\
& LM-SC$^{*}$ & 61.16\,($\pm$3.31) & 60.99\,($\pm$3.28) & 54.78\,($\pm$1.93) & 52.89\,($\pm$2.63) & 77.94\,($\pm$0.97) & 77.61\,($\pm$0.92)\\
& GFSC$^{*}$ & 52.77\,($\pm$0.38) & 52.01\,($\pm$0.56) & 54.426\,($\pm$2.47) & 53.15\,($\pm$2.70) & 75.96\,($\pm$1.50) & 76.14\,($\pm$1.32)\\
\midrule[1pt]
\multirow{7}{*}{\textbf{Text-Image}}
& MFN   & 55.86\,($\pm$1.66) & 52.81\,($\pm$1.45) & 50.91\,($\pm$2.86) & 49.20\,($\pm$3.05) & 78.98\,($\pm$1.60) & 78.28\,($\pm$2.10)\\
& CLMLF  & 56.97\,($\pm$2.08) & 52.04\,($\pm$2.35) & 49.63\,($\pm$2.40) & 45.72\,($\pm$2.17) & 74.33\,($\pm$2.85) & 72.51\,($\pm$1.95)\\
& TomBERT  & 55.95\,($\pm$5.17) & 43.248\,($\pm$0.06) & 47.47\,($\pm$2.26) & 36.93\,($\pm$5.89) & 72.34\,($\pm$2.37) & 70.55\,($\pm$3.04)\\
& EF-CapTrBERT  & 57.81\,($\pm$1.45) & 42.72\,($\pm$1.00) & 47.41\,($\pm$1.01) & 33.58\,($\pm$3.58) & -- & --\\
& KEF & 57.58\,($\pm$2.04) & 43.09\,($\pm$0.25) & 45.74\,($\pm$0.78) & 31.29\,($\pm$2.39) & -- & -- \\
& FITE & 58.42\,($\pm$0.18) & 43.29\,($\pm$0.11) & 46.20\,($\pm$0.52) & 29.97\,($\pm$0.70) & -- & -- \\
& VLP-MABSA & 53.36\,($\pm$1.07) & 43.23\,($\pm$3.75) & 55.32\,($\pm$3.39) & 48.96\,($\pm$1.26) & -- & -- \\
\midrule[1pt]
\multirow{8}{*}{\textbf{Text-Image}}
& PVLM$^{*}$  & 59.25\,($\pm$2.02) & 54.45\,($\pm$3.33) & 54.28\,($\pm$3.17) & 51.02\,($\pm$5.24) & 77.94\,($\pm$1.25) & 77.85\,($\pm$1.09)\\
& UP-MPF$^{*}$ & 61.56\,($\pm$2.43) & 60.16\,($\pm$2.54) & 54.93\,($\pm$2.22) & 51.87\,($\pm$4.08) & 77.75\,($\pm$2.14) & 77.84\,($\pm$1.93)\\
& MultiPoint(${\mathcal{P}^{f[1-4]}_{m}}$)   & 65.15\,($\pm$0.88) & 64.34\,($\pm$1.02) & 60.31\,($\pm$1.78) & 59.65\,($\pm$1.67) & 83.72\,($\pm$0.84) & 83.53\,($\pm$0.84)\\
& MultiPoint(${\mathcal{P}^{f[2-4]}_{m}}$)   & 66.23\,($\pm$0.83) & 65.59\,($\pm$1.09) & 60.18\,($\pm$1.86) & 59.41\,($\pm$1.77) & 82.73\,($\pm$1.05) & 82.53\,($\pm$1.04)\\
& MultiPoint(${\mathcal{P}^{f[3-4]}_{m}}$)   & \textbf{67.33\,($\pm$1.07)} & \textbf{66.61\,($\pm$1.36)} & \textbf{61.88\,($\pm$2.56)} & \textbf{61.23\,($\pm$2.58)} & \textbf{84.05\,($\pm$0.77)} & \textbf{83.86\,($\pm$0.86)}\\
& MultiPoint(${\mathcal{P}^{f[1-2]}_{m}}$) & 65.48\,($\pm$0.99) & 64.99\,($\pm$0.90) & 56.89\,($\pm$1.04) & 56.14\,($\pm$1.27) & 81.55\,($\pm$0.89) & 81.09\,($\pm$0.95) \\
& MultiPoint(${\mathcal{P}^{f[1-3]}_{m}}$) & 65.98\,($\pm$1.86) & 65.65\,($\pm$1.55) & 58.82\,($\pm$1.95) & 58.05\,($\pm$2.35) & 81.90\,($\pm$1.47) & 81.76\,($\pm$1.43) \\
& MultiPoint(${\mathcal{P}^{f[2-3]}_{m}}$) & 66.31\,($\pm$0.81) & 66.06\,($\pm$0.84) & 58.51\,($\pm$2.31) & 58.22\,($\pm$2.28) & 82.05\,($\pm$0.99) & 81.82\,($\pm$0.96)\\
\bottomrule[1pt]
\end{tabular}
\label{fine-grained results}
\end{center}
\vspace{-1em}
\end{table*}
\vspace{-1em}
\begin{table*}[t] \small
\begin{center}
\caption{Ablation experimental results about on Acc metric on six datasets. 
}
\vspace{-1em}
\renewcommand{\arraystretch}{1} 
\begin{tabular}{
p{2.5cm}< \centering|
p{2.2cm}< \centering| p{2.3cm}< \centering| p{1.6cm}< \centering| 
p{2.2cm}< \centering| p{2.2cm}< \centering| p{1.6cm}< \centering}
\toprule[1pt]
\textbf{Model} & \textbf{MVSA-Single} & \textbf{MVSA-Multiple} & \textbf{TumEmo} & \textbf{Twitter-2015} & \textbf{Twitter-2017} & \textbf{MASAD} \\
\cline{1-7}
w/o Image &  65.77\,($\pm$2.21) &  66.83\,($\pm$1.01) & 56.37\,($\pm$0.42) &  63.22\,($\pm$1.50) &  60.26\,($\pm$2.39) &  79.46\,($\pm$1.45)  \\
w/o Caption & 66.41\,($\pm$1.62) &  67.55\,($\pm$1.09) & 56.38\,($\pm$0.56) &  66.788\,($\pm$1.36) &  61.12\,($\pm$2.48) &  79.72\,($\pm$1.94)  \\
w/o MD &  69.56\,($\pm$1.89) & 67.86\,($\pm$0.97) & 57.88\,($\pm$0.47) &  64.77\,($\pm$1.56) &  61.28\,($\pm$2.72) &  82.57\,($\pm$1.08)  \\
w/ MultiPoint(${\mathcal{P}^{1}_{m}}$) & 67.62\,($\pm$1.77) & 66.09\,($\pm$1.75) & 56.94\,($\pm$0.99) & 63.72\,($\pm$1.39) & 57.62\,($\pm$1.57)  & 80.37\,($\pm$1.26) \\
w/ MultiPoint(${\mathcal{P}^{2}_{m}}$)   & 67.52\,($\pm$1.88) & \textit{67.25\,($\pm$1.63)} & 56.75\,($\pm$0.56) & 65.42\,($\pm$1.49) & 54.78\,($\pm$1.84) & 80.13\,($\pm$2.32) \\
w/ MultiPoint(${\mathcal{P}^{3}_{m}}$)   & \textit{68.84\,($\pm$2.38)} & 66.69\,($\pm$0.59) & 57.06\,($\pm$0.70) & \textit{66.25\,($\pm$1.05)} &  58.56($\pm$1.70) & 80.55\,($\pm$1.74) \\
w/ MultiPoint(${\mathcal{P}^{4}_{m}}$)   & 68.59\,($\pm$2.26) & 66.65\,($\pm$0.97) & \textit{57.19\,($\pm$0.59)} & 64.22\,($\pm$2.96) & \textit{60.52\,($\pm$4.11)} & \textit{82.33\,($\pm$1.06)} \\
w/ Average Fusion &  69.71\,($\pm$1.24) &  68.22\,($\pm$1.21) & 58.04\,($\pm$0.55) &  67.18\,($\pm$0.63) &  60.10\,($\pm$2.51) &  83.76\,($\pm$1.29)  \\
MultiPoint & \textbf{69.95\,($\pm$2.47)} & \textbf{68.27\,($\pm$1.15)} & \textbf{58.05\,($\pm$0.53)} & \textbf{67.33\,($\pm$1.07)} & \textbf{61.88\,($\pm$2.56)} & \textbf{84.05\,($\pm$0.77)} \\
\bottomrule[1pt]
\end{tabular}
\label{ablation results}
\end{center}
\vspace{-1em}
\end{table*}
\begin{table*}[h] \small
\begin{center}
\caption{Experimental results on Acc metric on few-shot datasets with the same amount of data for each category. 
The symbol $\nabla$ denotes the decrease in performance compared to our few-shot datasets based on CDS. 
}
\vspace{-1em}
\renewcommand{\arraystretch}{1} 
\begin{tabular}{
p{1.25cm}< \centering|
p{2.3cm}< \centering| p{2.4cm}< \centering| p{2.3cm}< \centering| 
p{2.3cm}< \centering| p{2.3cm}< \centering| p{2.3cm}< \centering}
\toprule[1pt]
\textbf{Model} & \textbf{MVSA-Single} & \textbf{MVSA-Multiple} & \textbf{TumEmo} & \textbf{Twitter-2015} & \textbf{Twitter-2017} & \textbf{MASAD} \\
\cline{1-7}
PVLM &  59.95\,($\pm$3.27) $\nabla$\textbf{6.99} & 59.18\,($\pm$3.21) $\nabla$\textbf{8.22}  & 52.67\,($\pm$0.95) $\nabla$\textbf{2.76}  & 51.17\,($\pm$6.78) $\nabla$\textbf{8.08} & 51.47\,($\pm$0.96) $\nabla$\textbf{2.81}  & 71.96\,($\pm$2.44) $\nabla$\textbf{5.98}  \\
UP-MPF &  61.75\,($\pm$3.82) $\nabla$\textbf{5.09} & 57.32\,($\pm$2.76) $\nabla$\textbf{10.03} & 51.44\,($\pm$1.78) $\nabla$\textbf{3.47} & 54.83\,($\pm$8.10) $\nabla$\textbf{6.73} & 53.21\,($\pm$2.46) $\nabla$\textbf{1.72} & 75.18\,($\pm$1.62) $\nabla$\textbf{2.57} \\
MultiPoint &  63.11\,($\pm$3.96) $\nabla$\textbf{6.84} & 61.14\,($\pm$1.66) $\nabla$\textbf{7.13}  & 55.15\,($\pm$0.47) $\nabla$\textbf{2.94} & 57.92\,($\pm$3.53) $\nabla$\textbf{9.41} & 58.46\,($\pm$2.75) $\nabla$\textbf{3.42} & 81.52\,($\pm$1.86) $\nabla$\textbf{2.53} \\
\bottomrule[1pt]
\end{tabular}
\label{Random experiments}
\end{center}
\vspace{-1em}
\end{table*}
\subsection{Experimental Results and Analysis}
Following \cite{DBLP:conf/icmcs/YuZ22, DBLP:conf/mm/YuZL22}, we report the results of our model and baselines on few-shot datasets with 1\% training data.
We introduce different combinations of multimodal prompts in MultiPoint from Table \ref{Templates}, such as [$\mathcal{P}^{c3}_{m}$, $\mathcal{P}^{c4}_{m}$] $\rightarrow$ $\mathcal{P}^{c[3-4]}_{m}$.
The performance comparison of our model (MultiPoint) with the baselines is shown in Table \ref{coarse-grained results} for coarse-grained MSA datasets and Table \ref{fine-grained results} for fine-grained MSA datasets. We make the following observations:

(1) Our model outperforms other robust models, including SOTA multimodal baselines (CLMLF and FITE), text-only prompt tuning models (PT, LM-BFF, LM-SC, and GFSC), and multimodal prompt tuning models (PVLM, UP-MPF). MultiPoint outperforms the existing SOTA few-shot multimodal model, UP-MPF, by more than 3-6\% on different datasets, especially for fine-grained datasets. This is due to our use of image prompts to bridge the gap between text and image modalities, introduction of multimodal demonstrations to improve the robustness of our model, and the utilization of probabilistic fusion modules to capture more practical information, including handcrafted prompts and learnable prompts.
(2) Our model yields varying results when using different combinations of prompts, and the combination of manual prompts and learnable prompts outperforms using only different manual prompts. 
(3) Most multimodal models trained on complete datasets outperform text-only models in the few-shot setting, indicating the importance of the image modality for sentiment analysis. However, multimodal models that perform very well on the full dataset perform poorly in the few-shot setting, like CLMLF, VLP-MABSA, and others, mainly due to overfitting on the few-shot data.
(4) Similar to previous studies, most prompt-based approaches (denoted with $*$) outperform state-of-the-art multimodal approaches (the second group) by a large margin, even using prompts to tune the model on the text-only modality.
(5) Prompt-based generative models for few-shot classification tasks perform poorly compared to cloze-based pre-trained masked language models. There is still much room for exploration using generative models to solve few-shot classification problems.
\subsection{Ablation Experiments}
We conduct ablation experiments on the MultiPoint model to demonstrate the effectiveness of its different modules, and the results are listed in Table \ref{ablation results}. Removing any of these modules affects the model's performance, indicating their significance in few-shot MSA. Here are our specific findings:
First, we remove the image modality (w/o Image), including image slots and captions, to verify the effectiveness of image information. The model's performance drops significantly, indicating that image modality is critical in few-shot MSA.
Second, we remove the image prompt (w/o Caption) and only apply image slots to the pre-trained language model. The model's performance drops drastically, suggesting that simply introducing image modalities into the pre-trained model fails to capture adequate image information due to the discrepancy of different modalities.
Third, we remove the Multimodal Demonstration (w/o MD) to verify the validity of multimodal demonstrations. The model's performance drops, indicating that multimodal demonstrations are effective in few-shot MSA.
Fourth, we utilize only one multimodal prompt, such as $\mathcal{P}^{1,2,3,4}_{m}$, to affirm the usefulness of our proposed multiple multimodal prompts and the probabilistic fusion module (PF). The results drop significantly across all datasets, suggesting that multiple multimodal prompts can furnish more informative few-shot sentiment analysis.
In the single-prompt setting, different datasets achieve the best results applying different prompts.
Note that the learnable prompt $\mathcal{P}_m^4$ achieves the best results on most datasets, such as TumEmo, Twitter-2015 and MASAD, followed by $\mathcal{P}_m^ 3$. These results show that the amount of information mined by different prompts is distinct, an observation further supported by the results for multiple prompts combinations in Table \ref{coarse-grained results} and Table \ref{fine-grained results}.
Finally, we replace the probabilistic fusion module with average fusion (w/ Average Fusion), i.e., averaging multiple logits from the model. The results on all datasets slightly decreased, indicating that the proposed probabilistic fusion module is effective.
\vspace{-1em}
\vspace{-1em}
\begin{figure}[H] 
  \centering 
\subfigure[MVSA-Single.]{
  \label{Fig3.sub.a}
  \includegraphics[scale = 0.25]{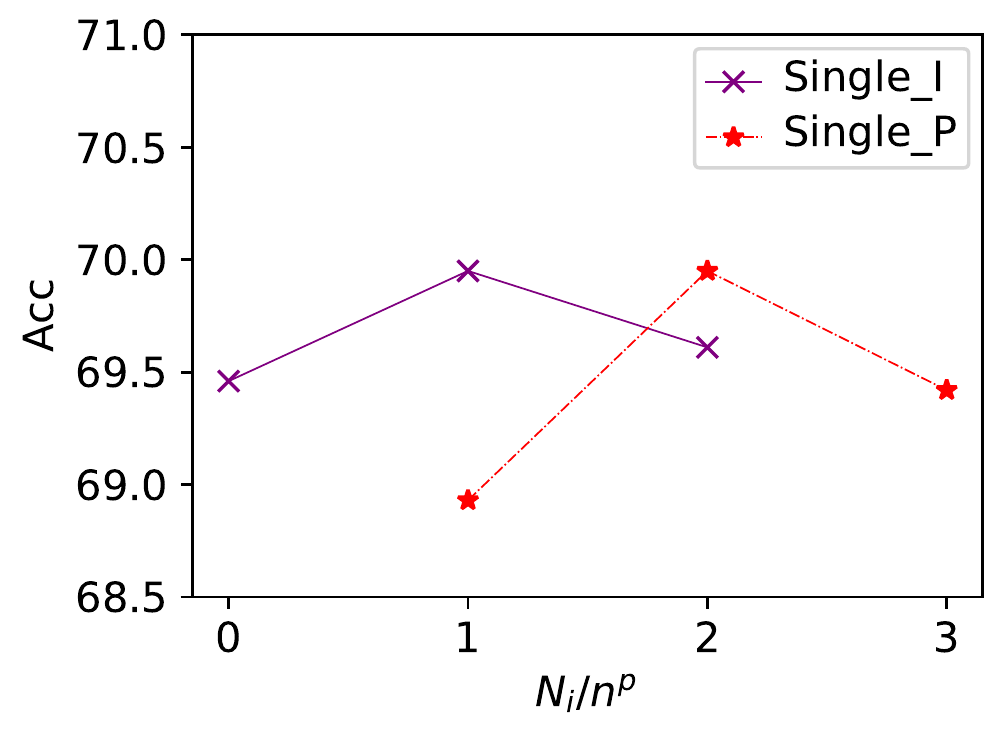}
  \vspace{-0.9em}
}
\subfigure[MVSA-Multiple.]{
  \label{Fig3.sub.b}
  \includegraphics[scale = 0.25]{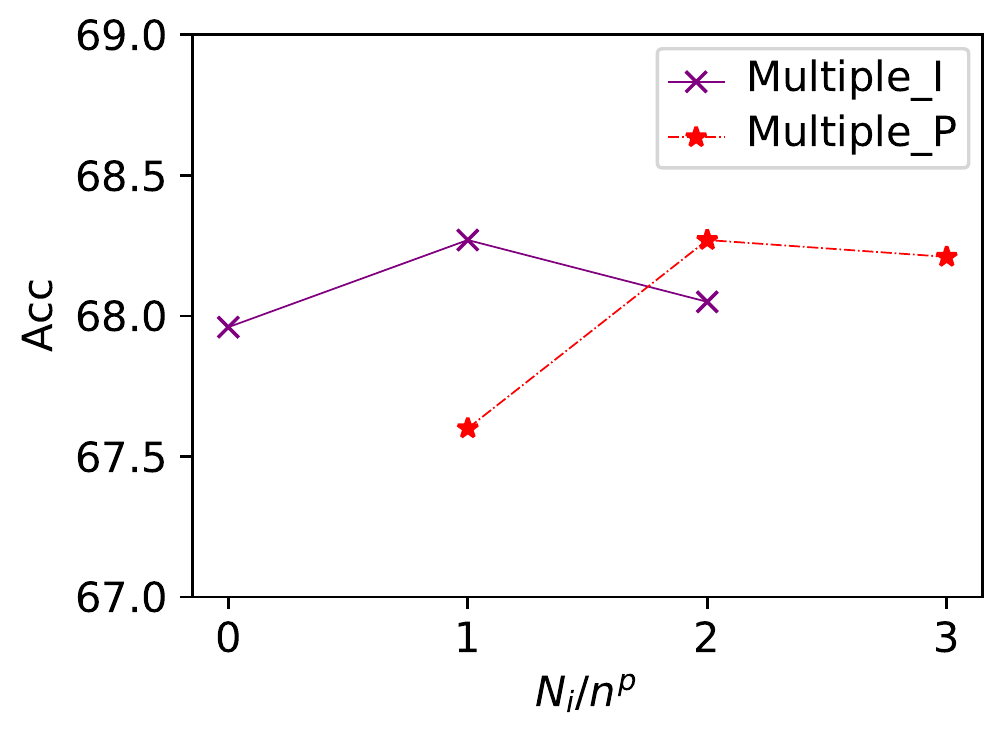}
  \vspace{-0.9em}
}
\subfigure[TumEmo.]{
  \label{Fig3.sub.c}
  \includegraphics[scale = 0.25]{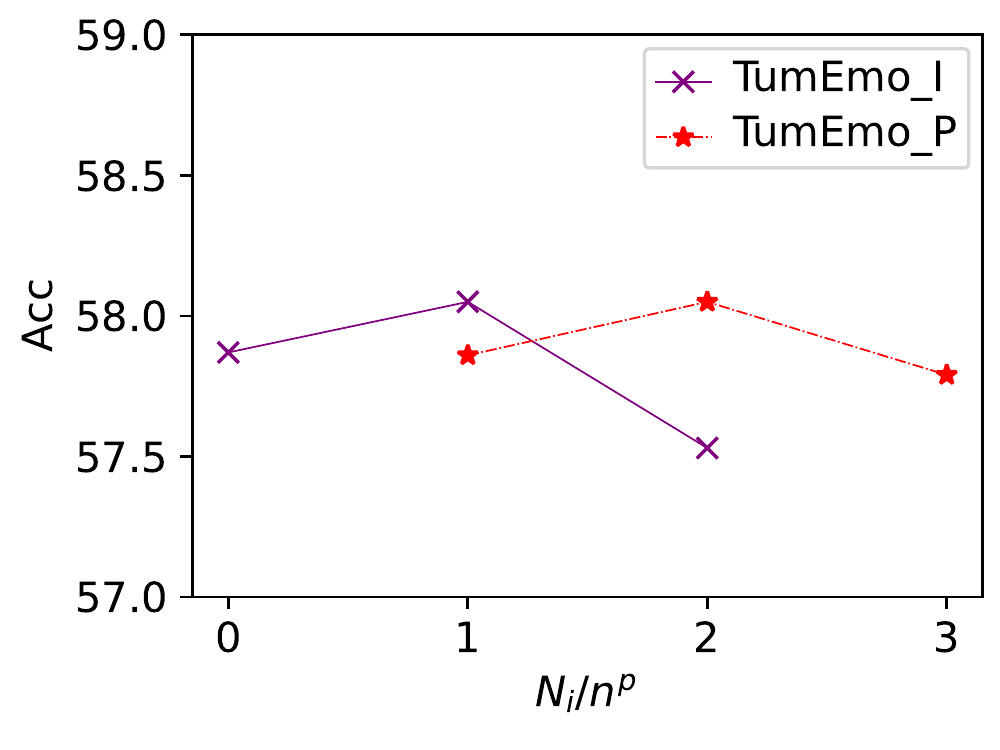}
  \vspace{-0.9em}
}
\subfigure[Twitter-2015.]{
  \label{Fig3.sub.d}
  \includegraphics[scale = 0.25]{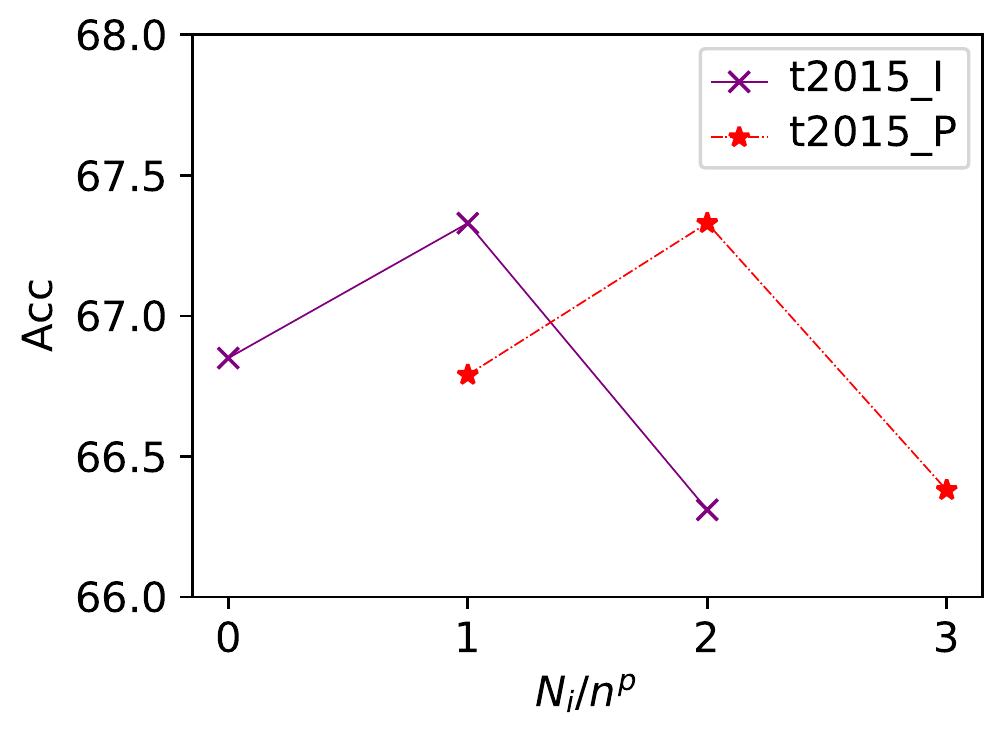}
  \vspace{-0.9em}
}
\subfigure[Twitter-2017.]{
  \label{Fig3.sub.e}
  \includegraphics[scale = 0.25]{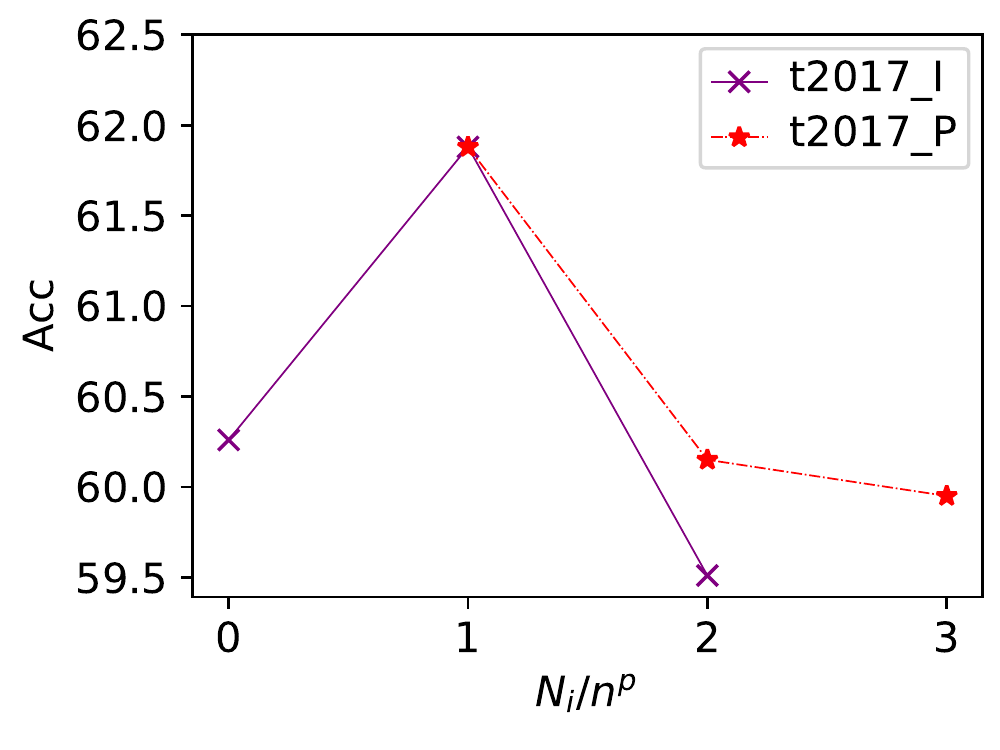}
  \vspace{-0.9em}
}
\subfigure[MASAD.]{
  \label{Fig3.sub.f}
  \includegraphics[scale = 0.25]{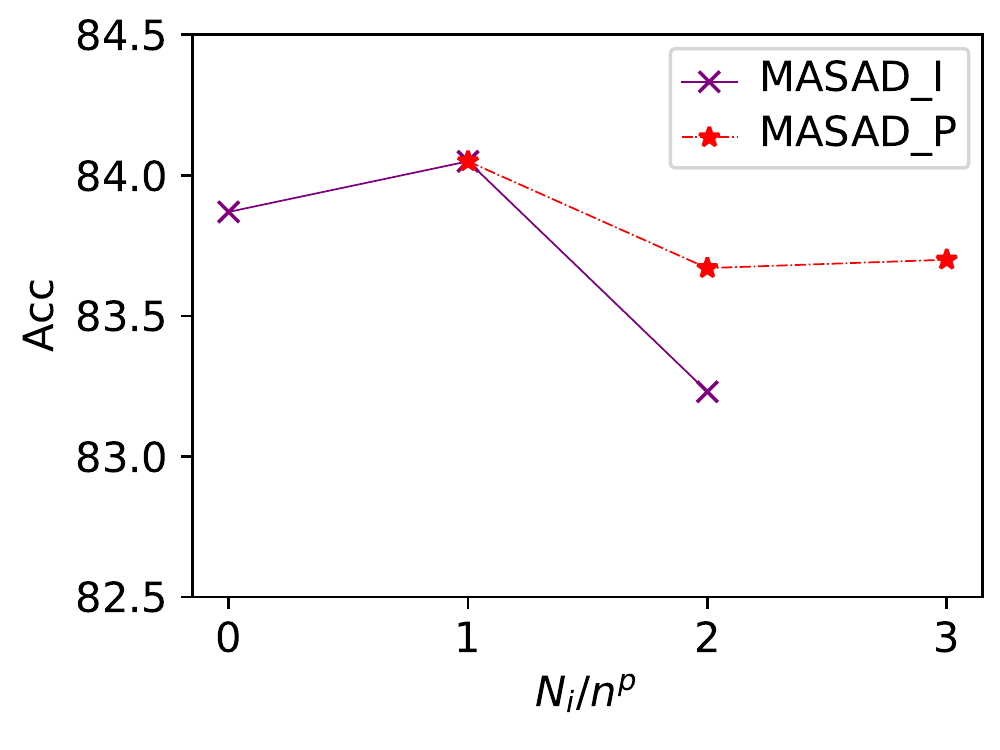}
  \vspace{-0.9em}
}
\vspace{-1em}
\caption{\textbf{Acc} comparisons of different Hyperparameters on different datasets, e.g., the number of image tokens, $N_i$, and the number of prompt tokens, $n^p$. $I$ means the image token, $P$ means the prompt token.} 
  \label{Tokens amount} %
  \vspace{-1em}
\end{figure}
\label{sec:experiments}
\subsection{Effect of Consistently Distributed Sampling}
We design diverse and comprehensive few-shot datasets based on CDS, as shown in Tables \ref{Dataset} and \ref{TumEmo}. Following the approach of \cite{DBLP:conf/icmcs/YuZ22, DBLP:conf/mm/YuZL22}, we sample the data to create \textbf{f}ew-\textbf{s}hot datasets with an \textbf{e}qual number of instances for each \textbf{s}entiment \textbf{c}ategory, \textbf{ESCFS}, while keeping the total amount of data consistent with few-shot datasets based on CDS.
We reproduce our model, MultiPoint, as well as the PVLM and UP-MPF models for the few-shot multimodal MSA task on these datasets, as reported in Table \ref{Random experiments}. We observe that the performance of each model on all datasets has decreased by 2.5-10\% when trained on ESCFS (indicated by the symbol $\nabla$), indicating the effectiveness of our few-shot datasets with consistent distribution. The CDS approach is particularly beneficial for smaller datasets, such as MVSA-Single, MVSA-Multiple, and Twitter-2015.
\subsection{Image Tokens and Prompt Tokens Amount}
In order to preserve adequate information from the image by NF-ResNet, we conduct experiments on all few-shot datasets under different settings of the hyperparameter $N_i$ in Eq. \ref{equ_3}, and the corresponding results are shown by solid purple lines in Figure \ref{Tokens amount}. We obtain the best performance for all datasets when $N_i = 1$. When $N_i$ is smaller, the image information is not fully utilized, while retaining more image features brings redundant information to the model.
We also leverage the continuous prompt tokens, $<PT>$ in $\mathcal{P}^4$, to mine knowledge from the pre-trained language model. We conduct hyperparameter experiments on the amount of prompt tokens, $n^p$, as the red dotted line shows in Figure \ref{Tokens amount}. Our model achieves the best performance on Twitter-2017 and MASAD when $n^p=1$, and on other datasets when $n^p=2$.
\vspace{-1em}
\section{Conclusion}
\label{sec:Conclusion}
In this paper, we first present a Consistently Distributed Sampling approach called CDS to construct the few-shot dataset with a category distribution
similar to that of the full dataset.
We further propose a novel approach to the few-shot MSA task, which is comprised of a Multimodal Probabilistic Fusion Prompts model with Multimodal Demonstrations (MultiPoint). Our model leverages a unified multimodal prompt, which combines image prompt and textual prompt, and dynamically selects multimodal demonstrations to improve model robustness. Additionally, we introduce a probabilistic fusion module to fuse multiple predictions from different multimodal prompts. Our extensive experiments on six datasets demonstrate the effectiveness of the CDS and the MultiPoint, outperforming state-of-the-art models on most datasets. In future work, we plan to explore more effective fusion approaches for different prompts to further improve the performance of few-shot multimodal sentiment analysis.

\section*{Acknowledgements}
Thanks to all co-authors for their hard work. The work is supported by National Natural Science Foundation of China (62172086, 62272092), Doctoral Research Innovation of Northeastern University (N2216004), and Chinese Scholarship Council.

\bibliographystyle{ACM-Reference-Format}
\bibliography{multipoint}

\end{document}